\documentclass[runningheads]{llncs}
\usepackage{graphicx}

\usepackage{tikz}
\usepackage{comment}
\usepackage{amsmath,amssymb} 
\usepackage{color}
\usepackage{hyperref}
\usepackage{tikz}
\usepackage{bm}
\usepackage{comment}
\usepackage{amsmath,amssymb} 
\usepackage{color}
\usepackage{wrapfig}
\usepackage{epsfig}
\usepackage{graphicx}
\usepackage{amsmath}
\usepackage{amssymb}
\usepackage{bm}
\usepackage{booktabs}
\usepackage{tabularx}
\usepackage{setspace}

\usepackage[accsupp]{axessibility}  
\usepackage{floatrow}
\newfloatcommand{capbtabbox}{table}[][\FBwidth]

\begin{document}
\pagestyle{headings}
\mainmatter
\def\ECCVSubNumber{2221}  

\title{Single Stage Virtual Try-on via Deformable Attention Flows} 

\titlerunning{Single Stage Virtual Try-on via Deformable Attention Flows}
%
\author{Shuai Bai \and
Huiling Zhou \and
Zhikang Li \and
Chang Zhou \and
Hongxia Yang}
\authorrunning{S. Bai et al.}
%
\institute{DAMO Academy, Alibaba Group, China\\
\email{\{baishuai.bs,zhule.zhl,zhikang.lzk,ericzhou.zc,yang.yhx\} 
@alibaba-inc.com}
}

\maketitle

\begin{abstract}
Virtual try-on aims to generate a photo-realistic fitting result given an in-shop garment and a reference person image. Existing methods usually build up multi-stage frameworks to deal with clothes warping and body blending respectively, or rely heavily on intermediate parser-based labels which may be noisy or even inaccurate. To solve the above challenges, we propose a single-stage try-on framework by developing a novel Deformable Attention Flow (DAFlow), which applies the deformable attention scheme to multi-flow estimation. With pose keypoints as the guidance only, the self- and cross-deformable attention flows are estimated for the reference person and the garment images, respectively. By sampling multiple flow fields, the feature-level and pixel-level information from different semantic areas are simultaneously extracted and merged through the attention mechanism. It enables clothes warping and body synthesizing at the same time which leads to photo-realistic results in an end-to-end manner. Extensive experiments on two try-on datasets demonstrate that our proposed method achieves state-of-the-art performance both qualitatively and quantitatively. Furthermore, additional experiments on the other two image editing tasks illustrate the versatility of our method for multi-view synthesis and image animation.

\keywords{Virtual try-on, Single stage, Deformable attention flows}
\end{abstract}

\section{Introduction}

Virtual try-on aims to generate a photo-realistic and reasonable try-on result based on an in-shop garment and a reference person image. In recent years, due to its potential applications in the fashion and e-commerce industries, it has received more and more attention. Recent methods \cite{ge2021parser,AGCPN,vitonhd,ge2021disentangled,zflow} have achieved considerable progress in generating realistic results and preserving details. However, this task is still challenging, especially under complex poses and large deformations, where most of existing methods are still suffering from misalignment or obvious artifacts.

\begin{figure}[t]
\centering{
\includegraphics[width= 12cm]{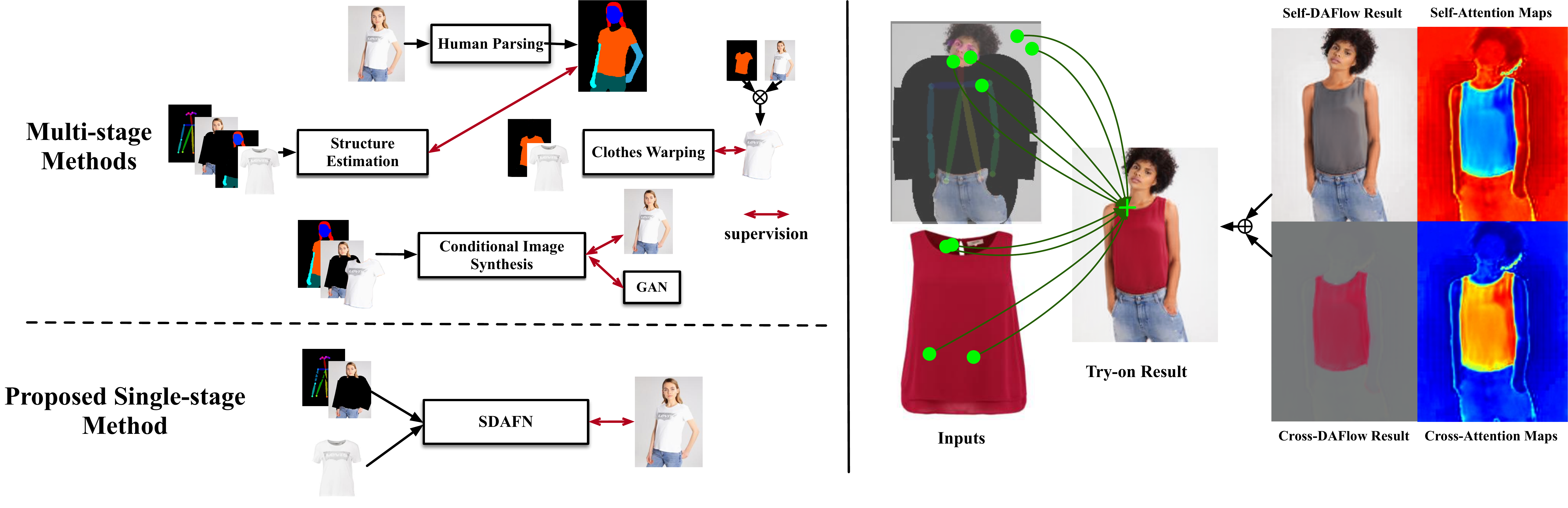}
   \caption{The comparison between multi-stage methods and our single-stage approach. With multiple flow fields sampled, different regions are associated. Attention maps learn the 3D priors to make the try-on more realistic without 3D information input.}
   \label{showcase}}
\end{figure}

Most prior try-on systems adopt a multi-stage approach \cite{han2018viton,CPVTON,AGCPN,ge2021disentangled} shown in Fig.~\ref{showcase}, including clothes warping, structure estimation, and image synthesis. Clothes warping is to align the garment to the target pose while preserving the texture details. Structure estimation predicts the segmentation map of the human body to guide the image synthesis. Given the warped clothing and the intermediate semantic labels, image synthesis is performed as a conditional generation task for pixel-level refinement. 

Clothes warping in recent works~\cite{han2019clothflow,AGCPN,ge2021parser,zflow} trains an extra flow network to warp clothes. Such flow operation retains the realistic texture but usually predicts the inaccurate structure. Some approaches~\cite{han2019clothflow,ge2021parser} limit the unsmooth deformation of flow by introducing additional constraints but fail in dealing with complex poses. To further improve the quality of the generated results, some works~\cite{AGCPN,zflow} introduce more prior knowledge from some external pre-trained models, such as human parsing~\cite{gong2017LIP}, densepose~\cite{guler2018densepose}, and 3D depth. However, the inaccurate intermediate predictions may lead to results with noticeable artifacts. In addition, although the generative adversarial networks (GAN)~\cite{gan} help preserve the sharp details of the generated images, they modify some attributes of clothes, such as color or style, which are not desirable for virtual try-on.

To address the above problems, we propose a Single-stage Deformable Attention Flow Network (SDAFN) to perform an end-to-end try-on task. we build a Deformable Attention Flow (DAFlow) module by applying a deformable attention scheme to multi-flow estimation. As shown in Fig.~\ref{showcase}, it estimates multiple flow fields from different semantic areas and then synchronously merges the feature-level and pixel-level information with the attention mechanism in cascade. This allows the extract deformable attention flows to not only warp clothes but also synthesize photo-realistic human torso and shadows at the same time. Then, we combine self- and cross-DAFlows to deal with the human body and the garment respectively, generating the fitting results in one single pass. In addition, our method only need pose keypoints as guidance. To our best knowledge, we are the first one-stage pure flow-based virtual try-on method.

The main contributions of this paper can be summarized as follows:

\begin{itemize}
    \item We propose a single-stage virtual try-on framework, which applies self- and cross-DAFlows to deal with the reference person and garment images in parallel and generate realistic fitting results in an end-to-end manner.
    \item We propose a novel deformable attention flow module to estimate the reasonable structure while retaining the vivid texture in cascade. It works well even with a large misalignment between the clothes and the person and can be applied to a variety of resolutions.
    \item Extensive experiments show that our proposed method not only achieves superior performance on virtual try-on, but also can be extended to other image editing tasks, such as multi-view synthesis and image animation.
\end{itemize}

\section{Related Work}

\subsection{Virtual Try-on.}
The study on the virtual try-on task mainly consists of 3D model-based approaches ~\cite{bhatnagar2019multi,mir2020learning,bertiche2020cloth3d,lahner2018deepwrinkles} and 2D image-based approaches~\cite{han2018viton,CPVTON,li2021toward,yu2019vtnfp}. Recently, 2D methods have attracted more and more attention because of highly accessible data and photo-realistic results. VITON~\cite{han2018viton} uses a two-stage coarse-to-fine strategy to generate a clothed person. It first estimates the rough human body shape, then warps clothes and refines the details of the clothed person according to the shape. To improve the accuracy, SwapNet~\cite{raj2018swapnet} and VTNFP~\cite{yu2019vtnfp} adopt semantic segmentation as guidance. ACGPN~\cite{AGCPN} introduces an additional stage to predict the semantic layout of the reference image. DCTON~\cite{ge2021disentangled}  and ZFlow~\cite{zflow} add more accurate descriptor, like densepose ~\cite{guler2018densepose} or UV~\cite{feng2018joint} projection. VITON-HD~\cite{vitonhd} improves the performance of the conditional GAN structure on high-resolution images.  Although the generated images get better quality, the pipeline is becoming more complex and relies on more external information. It results in a reduction in efficiency and obvious artifacts caused by the inaccurate intermediate labels. Recently WUTON~\cite{issenhuth2020not} and PFAFN~\cite{ge2021parser} adopt the "teacher-tutor-student" scheme to get rid of the extra information at inference time and achieve good performance. This proves that the network has the ability to perform try-on without heavy dependency. Our proposed method is able to obtain realistic fitting performance with only pose information as guidance. 

\subsection{Spatial Transform Module.}
The spatial transform module~\cite{jaderberg2015spatial,zhou2016view,ren2020deep,sun2018pwc,yu2020motion} is widely applied in optical flow estimation, image transformation and object recognition tasks. In virtual try-on, the spatial transform module is mainly used to wrap clothes to match the posture of the person. VITON~\cite{han2018viton} exploits a Thin-Plate Spline (TPS)~\cite{duchon1977splines} based warping method to deform the in-shop clothes to the refined result with a composition mask. CP-VTON~\cite{CPVTON} uses a neural network to learn the transformation parameters of TPS warping rather than using image descriptors. ClothFlow~\cite{han2019clothflow} introduces denser flow predictions through a cascade scheme, which makes deformation has a high dimension of freedom. 
However, dense flows usually present unappealing artifacts. To avoid this problem, ClothFlow uses flow variation regularization to enforce smoothness. PFAPN~\cite{ge2021parser} adds a second-order smooth constraint to encourage the co-linearity of neighboring appearance flows. These constraints are proposed to make the garment smoother after deformation, but they still face the challenge when there exists a huge discrepancy between the target clothes and the original clothes on the model.

\subsection{Efficient Attention Mechanism.} The attention mechanism~\cite{vaswani2017attention} is widely used in transformer models, benefiting from its long-distance association ability, and achieves good performance on image segmentation, object detection, and image generation. However, with the increase in feature resolution, dense attention will make a huge computational cost. This makes the attention mechanism usually used in low-resolution features~\cite{ren2021combining}. Recently sparse attention has been introduced into detection and recognition tasks. Common methods include pre-defined local attention patterns~\cite{liu2018generating,liu2021swin,qiu2019blockwise} or learned sparse attention~\cite{zhu2020deformable,wang2020linformer,kitaev2020reformer,tay2020sparse}. Combing the effectiveness of flow operation on preserving details and the capability of attention mechanism on estimating accurate structures, our proposed method naturally extend learned sparse attention to pixel-level image transformation.

\section{Our Approach}

In this section, we introduce the overall framework of our proposed single-stage virtual try-on model called SDAFN. Then, we describe the core DAFlow module as well as the training loss design in detail. 

\begin{figure}[t]
\centering{
\includegraphics[width= 12cm]{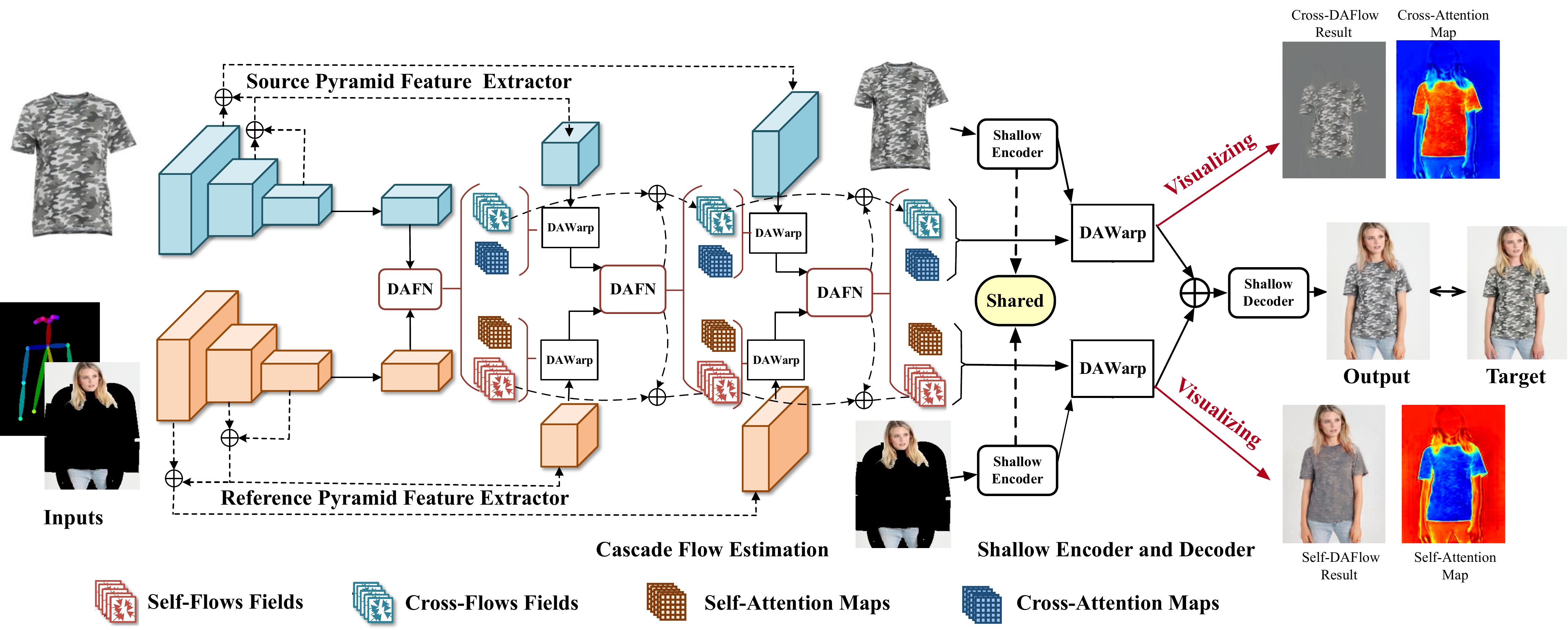}
   \caption{The overall framework of our SDAFN. The garment, person and pose images are firstly fed into the unshared pyramid feature extractors. Then, both self- and cross- deformable attention flows are estimated in cascade. The final try-on result is obtained by applying shallow encoder and decoder together with the learned flows.}
   \label{framework}}
\end{figure}

\subsection{Architecture\label{sec:arc}}
Different from the previous methods who apply the clothes warping and the conditional generation in different stages, we use self- and cross-deformable attention flows to obtain the try-on results directly. In addition, our framework only relies on the pose keypoints as guidance. Specifically, as shown in Fig.~\ref{framework}, our model is composed of three parts: pyramid feature extraction, cascade flow estimation, and shallow encoder-decoder generation.

\textbf{Pyramid feature extraction.} 
Our model has two pyramid feature extractors, including a reference branch and a source branch. The reference branch takes the concatenation of the person and pose images as input, where the upper body of the person is masked. The garment image is fed into the source branch. The two feature extractors have the same Feature Pyramid Network (FPN)~\cite{fpn} structure with unshared parameters. The FPN network consists of $N$ encoding layers where each layer has a downsampling convolution with a stride of 2 followed by two residual blocks~\cite{resnet}. 

\textbf{Cascade flow estimation.} It is difficult to directly and accurately predict large deformations, especially for the virtual try-on application, where the target and source images are not domain-unified. Hence, similar to the recent methods \cite{han2019clothflow,ge2021parser}, we adopt the cascade flow estimation from coarse to fine.

Given the hierarchical reference and source features $\{\bm{x}_r^n,\bm{x}_s^n\}_{n=1}^N$ extracted by the pyramid feature extractors, as illustrated in Fig.~\ref{framework}, two types of flows and attention maps are estimated. The first is the self-flow fields and self-attention maps $(\bm{o}_{daf,r}^n,\bm{a}_r^n)$ from the reference branch. The second is the cross-flow fields and cross-attention maps $(\bm{o}_{daf,s}^n,\bm{a}_s^n)$ from the interaction of the both branches.
The features $(\bm{x}_r^1,\bm{x}_s^1)$ at the lowest resolution are fed into the proposed Deformable Attention Flow Network (DAFN) to predict the initial flow fields $(\bm{o}_{daf,r}^1, \bm{o}_{daf,s}^1)$ and attention maps $(\bm{a}_r^1, \bm{a}_s^1)$. Then they will be refined and updated in cascade. Specifically, the reference and source feature maps $(\bm{x}_r^n,\bm{x}_s^n),n \in [2,N]$ are first transformed 
by the self- and cross-DAFlows  $(\bm{o}_{daf,r}^{n-1}, \bm{o}_{daf,s}^{n-1}, \bm{a}_r^{n-1}, \bm{a}_s^{n-1})$ of the previous scale, called Deformable Attention Warping (DAWarp). Then, the transformed features $(\hat{\bm{x}}_r^n,\hat{\bm{x}}_s^n)$ are applied to predict the residual flows and the new attention maps in which finer flows are obtained. The above process continues until $n = N$. In our implementation, $N$ is set to 5, and the process shown in Fig.~\ref{framework} is illustrated with $N =3$. Both DAFN and DAWarp modules are described in Sec.~\ref{sec:daf} in details.

\textbf{Shallow encoder-decoder generation.} The shallow encoder projects the images from RGB to high dimensional space. With the final estimated flows and attention maps, the reference person and garment images in high dimensional feature space are transformed and merged with DAWarp. Then the merged result is fed into a shallow decoder to obtain the final try-on result. The shallow encoder and decoder both have two convolution layers without downsampling. In particular, both garment and reference person images share the same encoder and decoder. It is found in our experiment that the shallow encoder-decoder structure is a simple and effective way to enhance the representation capability at the pixel level and improve the flow generation quality.

\begin{figure}[t]
\centering
\includegraphics[width= 12cm]{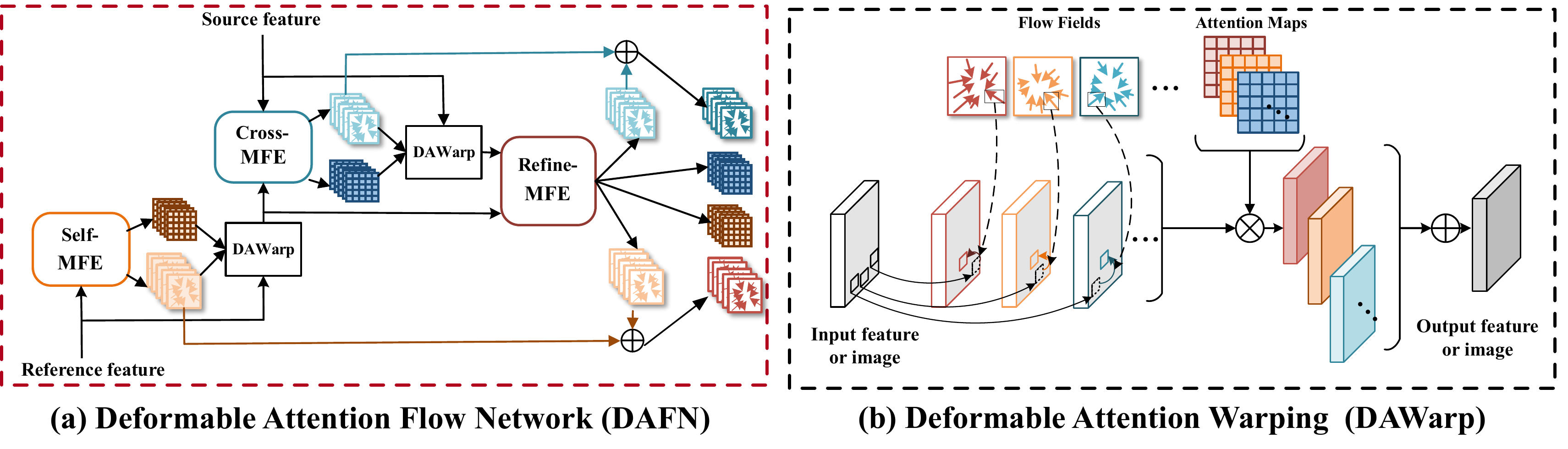}
   \caption{Illustrations of DAFN network and DAWarp operation.
   \label{DAF}}

\end{figure}

\subsection{Deformable Attention Flows\label{sec:daf}}

\textbf{Revisiting the conventional flow field.} 
The flow field estimation~\cite{han2019clothflow,zflow,ge2021parser} is extensively used in the cloth warping of virtual try-on.  It takes the reference (e.g., semantic segmentation, pose keypoints or depth map) feature $\bm{x}_r \in \mathbb{R}^{C\times H\times W}$ and the source (e.g., clothes) feature $\bm{x_s} \in \mathbb{R}^{C\times H\times W}$ as inputs, and only estimates one 2D coordinate offset for each sampling position. This process of predicting offset map $\bm{o} \in \mathbb{R}^{2\times H\times W}$ can be written as:
\begin{equation}
\bm{o} = F([\bm{x}_r,\bm{x}_s]), 
\end{equation}
where $C$ is the feature dimension, $H$ and $W$ denote the height and width of  feature map.  $F$ indicates flow field estimator network. Accordingly, the warped feature at a 2D reference location $p$ is calculated as :
\begin{equation}
\mathcal{W}(\bm{x}_s,\bm{o})_p = \bm{x}_s(p+ \bm{o}_p),
\end{equation}
where $\mathcal{W}(\cdot)$ denotes the warp operation, $p + \bm{o}_p$ is the estimated sampling position. The value $\bm{x}_s(p+\bm{o}_p)$ of each sampling point
is computed by bilinear interpolation from the nearby grid
points on the feature map. 

The flow operation directly samples from images, retaining the realistic textures. However, structural information is often associated with multiple locations. For example, the shape of the clothes on the human body is determined by the posture, body shape, and the type of garment. Hence, only one flow field cannot estimate accurate structure, so most existing methods only apply the flow to warp the garment in virtual try-on. To mitigate this weakness, we propose a deformable attention flow (DAFlow) module, which preserves detailed textures and accurately estimates structures. The DAFlow module consists of the Multiple Flow field Estimator (MFE) and the DAWarp operation.

\textbf{Multiple flow field estimator.} Inspired by deformable attention \cite{zhu2020deformable}, our MFE predicts the fixed number $K$ of sampling points, regardless of the spatial size of the feature maps. In contrast to the conventional single flow field $\bm{o} \in \mathbb{R}^{2\times H\times W}$, our MFE forecasts multiple flow fields  $\bm{o}_{daf} \in \mathbb{R}^{2K\times H\times W}$ and attention weight maps $\bm{a} \in \mathbb{R}^{K\times H\times W}$:
\begin{equation}
\bm{o}_{daf}, \bm{a} = F([\bm{x}_r,\bm{x}_s]), 
\end{equation}
where $K$ is the sampled key number. In the implementation, both $\bm{o}_{daf} $ and $\bm{a}$ are obtained via the same ConvNet but different channels.

\textbf{Deformable Attention Warping.} 
As shown in the \textbf{(b)} of Fig.~\ref{DAF}, each location is associated with features from multiple locations. Besides, DAFlows learn a variety of possible flows. The multiple flow fields and the attention operation are simultaneously applied to the features and images. In terms of the feature, the source or reference features $\bm{x} \in \mathbb{R}^{C\times H\times W}$ are effectively integrated into the desired target position as:
\begin{equation}
\mathcal{W}_{DAF}(\bm{x},\bm{o}_{daf},\bm{a})_p= \sum_{k=1}^{K}\frac{exp(\bm{a}_{pk})}{\sum_{i=1}^{K} exp(\bm{a}_{pi})} \bm{x}(p+ \bm{o}_{daf,pk}),
\end{equation}
where $(p+ \bm{o}_{daf,pk})$ is the $k_{th}$ sampling position for the reference location $p$, and  $\mathcal{W}_{DAF}(\bm{x}_s,\bm{o}_{daf},\bm{a}) $ denotes the warped feature with DAFlows. As for the image level, merging multiple warped images has more generation possibilities and is recombined into new images with reasonable structures and realistic textures. As illustrated in Fig.~\ref{att}, the neck and arm parts are realistically generated, which are not in the original image. Additionally, the warping operation applies bilinear interpolation, enabling the estimated offsets optimized with back-propagation.

\subsection{Deformable Attention Flow Network.}
As illustrated in Sec.~\ref{sec:arc}, we adopt cascade flow estimation to predict the deformation and attention from coarse to fine. Given the reference and source feature maps $(\bm{x}_r^n,\bm{x}_s^n),n\in [2,N]$, the feature maps are transformed to $(\bm{\hat{x}}_r^n,\bm{\hat{x}}_s^n)$ in accordance with the flows  $(\bm{o}_{daf,r}^{n-1}, \bm{o}_{daf,s}^{n-1})$ and attention $(\bm{a}^{n-1}_r,\bm{a}^{n-1}_s)$ from the previous scale:
\begin{equation}
\hat{\bm{x}}_{s/r}^n = \mathcal{W}_{DAF}(\bm{x}^{n}_{s/r},\mathcal{U}(\bm{o}_{daf,{s/r}}^{n-1}),\mathcal{U}(\bm{a}_{s/r}^{n-1})),
\end{equation}
where $\mathcal{U}(\cdot)$ denotes the bilinear samping with the scale factor of 2.

 For each scale, DAFN is utilized to estimate the DAFlows. The \textbf{(a)} of Fig.~\ref{DAF} illustrates that the DAFN consists of three flow estimators $(F^n_r,  F^n_s, F^n_{m})$, namely, Self-MFE, Cross-MFE and Refine-MFE. 
Firstly, the transformed reference feature $\bm{\hat{x}}_r^n$ is fed into $F^n_r$ to estimate the initial residual self-flows and the self-attention maps $(\Delta\bm{\dot{o}}_{daf,r}^n,\bm{\dot{a}}_r^n)$:
\begin{equation}
\Delta\bm{\dot{o}}_r^n,\bm{\dot{a}}_r^n = F^n_r(\bm{\hat{x}}^n_r).
\end{equation}
Secondly, the reference feature is transformed with $(\Delta\bm{\dot{o}}_r^n,\bm{\dot{a}}_r^n)$, and concatenated with source feature $\bm{\hat{x}}_s^n$ to forecast the residual cross-flow fields and cross-attention maps $(\Delta\bm{\dot{o}}_{daf,s}^n,\bm{\dot{a}}_{s}^n)$: 
\begin{equation}
\Delta\bm{\dot{o}}_{daf,s}^n,\bm{\dot{a}}_{s}^n = F^n_s([\bm{\hat{x}}^n_s,\mathcal{W}_{DAF}(\bm{\hat{x}}^n_r,\Delta\bm{\dot{o}}_r^n,\bm{\dot{a}}_r^n)]).
\end{equation}

Finally, the source feature is converted with $(\Delta\bm{\dot{o}}_{daf,s}^n,\bm{\dot{a}}_{s}^n)$, and combined with the newly transformed reference features to predict the refinements $(\Delta\bm{\ddot{o}}_{daf,r}^n,\Delta\bm{\ddot{o}}_{daf,s}^n)$ and the final attention maps $(\bm{a}_r^n,\bm{a}_s^n)$:
\begin{equation}
\Delta\bm{\ddot{o}}_{daf,s}^n, \Delta\bm{\ddot{o}}_{daf,r}^n ,\bm{a}_s^n,\bm{a}_r^n= F^n_{m}([\mathcal{W}_{DAF}(\bm{\hat{x}}^n_s,\Delta\bm{\dot{o}}_s^n,\bm{\dot{a}}_s^n),\mathcal{W}_{DAF}(\bm{\hat{x}}^n_r,\Delta\bm{\dot{o}}_r^n,\bm{\dot{a}}_r^n]).
\end{equation}
The final self- and cross-flow fields at the $n_{th}$ scale are:
\begin{equation}
\bm{o}_{daf,r}^n = \mathcal{U}(\bm{o}_{daf,r}^{n-1}) + \Delta \bm{\dot{o}}_{daf,r}^n+ \Delta \bm{\ddot{o}}_{daf,r}^n,
\end{equation}
\begin{equation}
\bm{o}_{daf,s}^n = \mathcal{U}(\bm{o}_{daf,s}^{n-1}) + \Delta \bm{\dot{o}}_{daf,s}^n+ \Delta \bm{\ddot{o}}_{daf,s}^n,
\end{equation}
and the output $\bm{a}_s^n,\bm{a}_r^n$ of Refine-MFE is applied as the final attention maps.

\subsection{Train Losses}
Discarding the dependence on intermediate parser-based labels, our model only uses the try-on result as supervision in an end-to-end manner. As for the loss functions, without any constraints or regularization on the flows, the model is directly optimized by comparing the similarity between generated results and ground truths in all scales. The similarity functions consist of the L1 loss, the perceptual loss~\cite{johnson2016perceptual}, and the style loss.
The L1 loss is formulated as:
\begin{equation}
\mathcal{L}_{L1} = ||\bm{I}_{out} - \bm{I}_{target} ||_1,
\end{equation}
where $||\cdot||_1$ indicates the $L1$ distance, and $\bm{I}_{out}$ and $\bm{I}_{target}$ represent the predicted image and target image. The perceptual loss proposed in~\cite{johnson2016perceptual} calculates the L1 distance of the feature maps extracted by the VGG-19~\cite{vgg} network:
\begin{equation}
\mathcal{L}_{prec} = \sum_{i=1}^5 ||\phi_i (\bm{I}_{out}) - \phi_i(\bm{I}_{taregt}) ||_1,
\end{equation} 
where $\phi_i(\bm{I}_{out})$ is the features of the $i_{th}$ layer. In addition, the style loss optimizes the statistical error between the feature maps:
\begin{equation}
\mathcal{L}_{style} = \sum_i ||G_i^\phi (\bm{I}_{out}) - G_i^\phi\bm{I}_{taregt}) ||_1,
\end{equation}
where $G_i^\phi$ denotes the Gram matrix of features. The overall loss is presented as:
\begin{equation}
\mathcal{L} = \sum_{n=1}^N(n+1)*(\lambda_{L1} \mathcal{L}_{L1}^n + \lambda_{prec} \mathcal{L}_{prec}^n + \lambda_{style} \mathcal{L}_{style}^n)
\end{equation}
where $\mathcal{L}^n$ is the loss of the $n_{th}$ scale, and the scale closer to the output is given a larger weight $n-1$.

\section{Experiments}

 \subsection{Datasets.} 
 We conduct experiments on VITON~\cite{han2018viton} and MPV~\cite{MPV} datasets. VITON dataset is commonly used in virtual try-on. it contains a training set of 14,221 image pairs and a testing set of 2,032 image pairs. Each pair has a front-view photo and an in-shop clothes image with the resolution $256 \times 192$. MPV dataset is a recently virtual try-on dataset with multiple views, containing 35,687 / 13,524 person/clothes images at $256 \times 192$ resolution where a test set of 4175 image pairs is split out. To fair comparison, following \cite{issenhuth2020not,ge2021parser}, the images tagged as back ones are removed since the target garment is only from the front.

\subsection{Implementation Details.} 
\textbf{Network structure.} The two encoders have the FPN~\cite{fpn} structure with five layers, each layer consists of a downsampling convolution with the stride of 2, followed by two residual blocks. The MFE flow estimator in each layer comprises ConvNets with four convolution layers, and the hidden dimensions are $[256,128,64,32]$. To obtain a large receptive field, the kernel size of the last three convolution layers is set to 7. the shallow encoder and decoder are ConvNets with two convolution layers without downsampling, and the hidden dimensions are $[32,64]$. The $K = 6$ in our experiment. Our approach only has a single stage, the computational efficiency is similar to the clothes warping stage in \cite{han2019clothflow,ge2021parser}. 

\textbf{Training details.} We adopt the same training parameters for the two datasets. All our experiments are conducted using Pytorch on Tesla V100 GPUs. The AdamW~\cite{loshchilov2017decoupled} optimizer is adopted with a batch size of 8. We train the model for 200 epochs where the initial learning rate is $5\times 10^{-5}$ and is reduced to 0.1 times the original every 50 epochs. We set the weight of the loss function $\lambda_{L1}=1, \lambda_{prec}=1, \lambda_{style}=100$.

\subsection{Qualitative Results.}
\begin{figure}[t]
\centering
\includegraphics[width= 12cm]{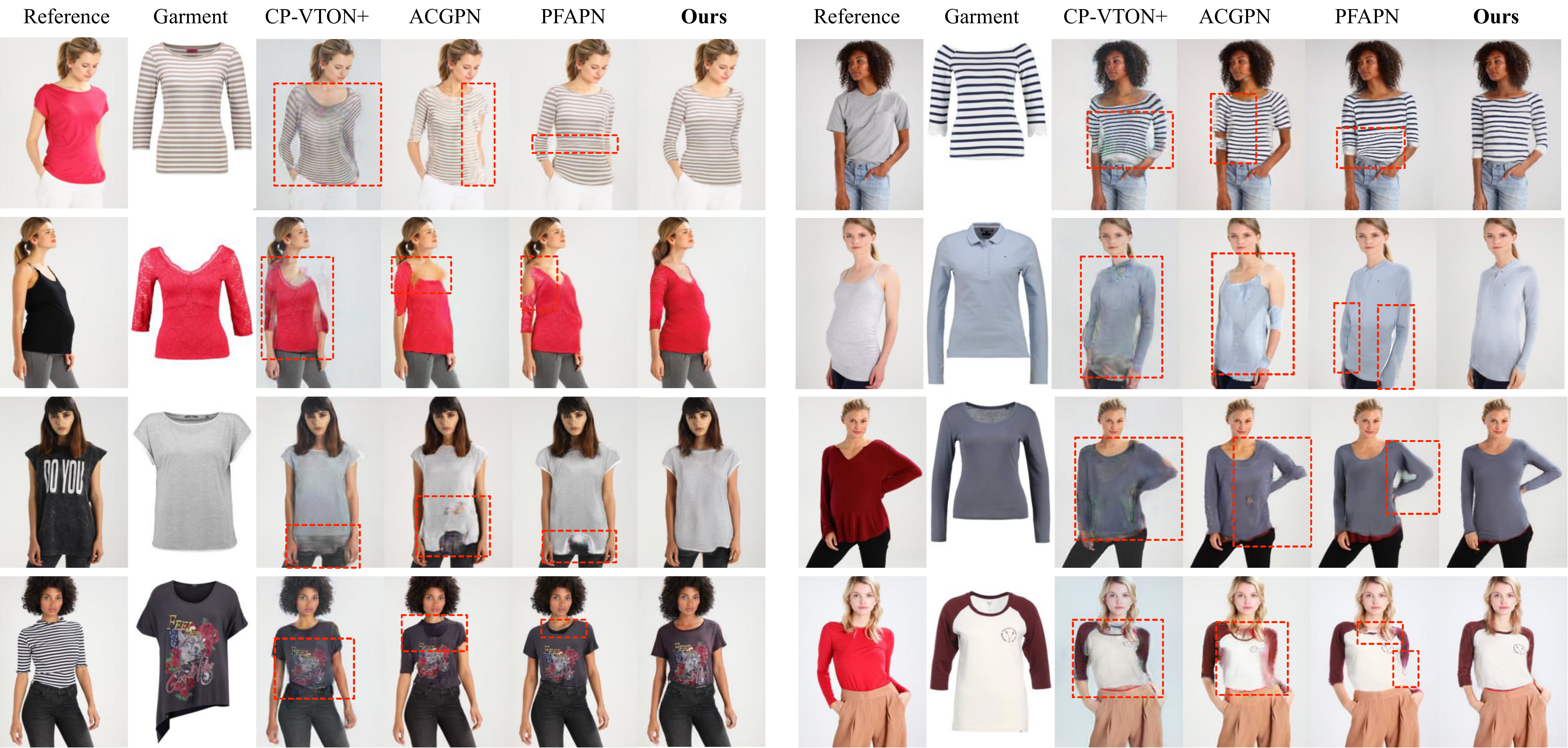}

   \caption{Visual comparison on VITON dataset. Different regions are marked in red.\label{viton}}

\end{figure}

\textbf{Results on VITON.} To qualitatively evaluate our method, we visually compare our method with three recently proposed virtual try-on works with available code implementations, including CP-VITON+~\cite{cpvtonplus}, ACGPN~\cite{AGCPN}, PFAPN\cite{ge2021parser}, as shown in Fig.~\ref{viton}. It shows that all comparing works are able to roughly align clothes with the target person pose. However, noticeable artifacts are observed from their results where there are complex poses or misalignment occurs between the target clothes and the person.

As shown in the first row of Fig.~\ref{viton}, baseline methods fail to preserve the striped pattern after warping, especially around the highly non-rigid body parts like the forearm and waist. The second row shows the results with the side view where baseline methods are not able to deal with large deformation in poses and lead to blur or incorrect fitting results. In comparison, our proposed method is capable of extracting accurate structural and textural information and performing reasonable warping even when there exists huge discrepancy (e.g. large poses or long-sleeve in target clothes while short-sleeve in reference image). The last two columns that parser-based methods like CP-VITON+~\cite{cpvtonplus} and ACGPN~\cite{AGCPN} are delicate to segmentation errors while learning warping flows. They do not always produce reasonable results for areas like necklines and lower-body parts. Even though PFAPN is a parser-free framework with less distortion in clothes warping, it cannot preserve or generate the body parts well which results in blurry arms and shoulders. Unlike them, our method clearly preserves the characteristics of both the target clothes as well as the body parts, benefiting from the proposed self- and cross-DAFlows.

\begin{figure}[t]
\centering
\includegraphics[width= 12cm]{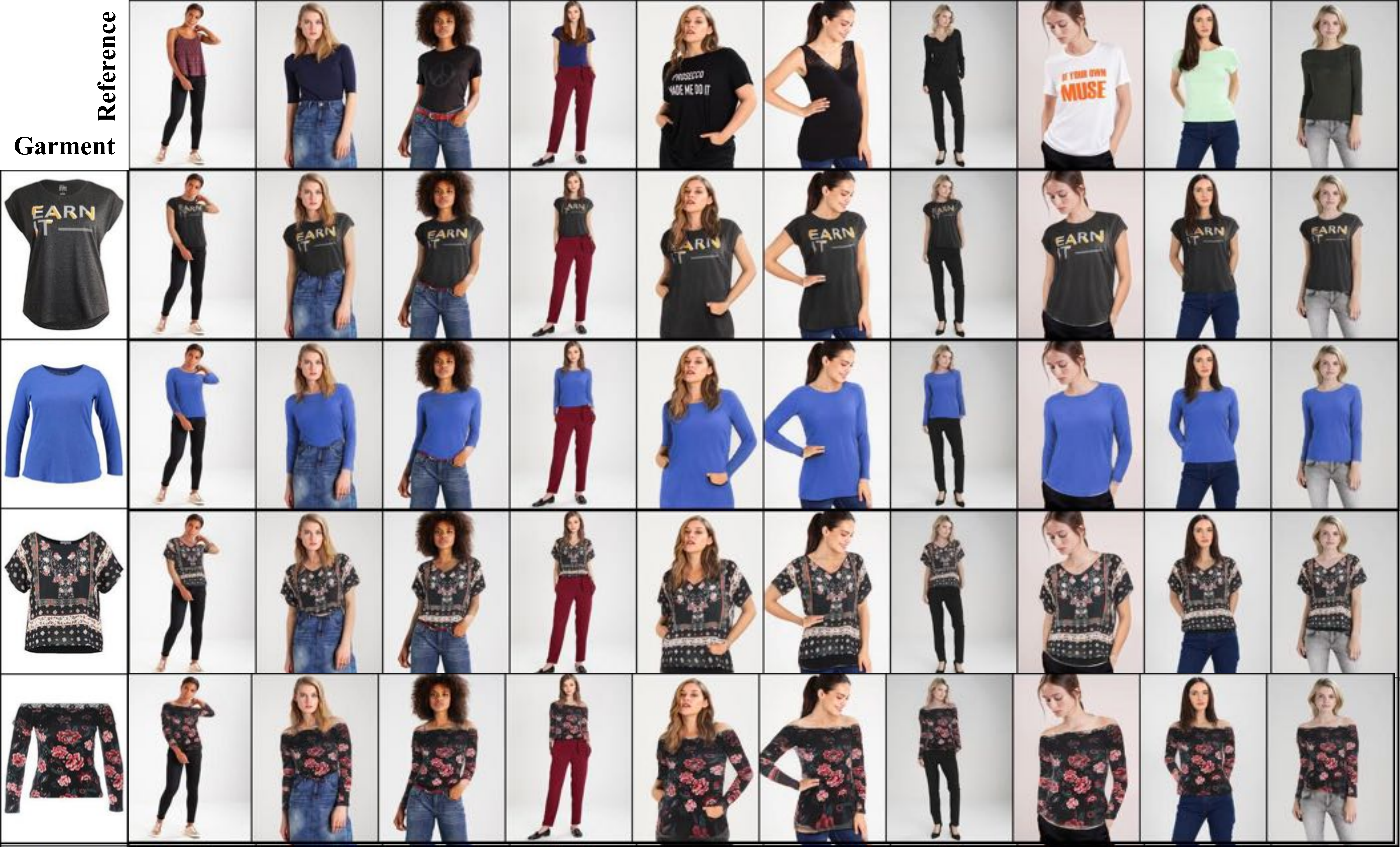}
   \caption{From diverse perspectives, our approach generates high-quality images and accurately preserves vivid attributes on MPV dataset.}\label{mpvvis}
\end{figure}

\textbf{Results on MPV.}
To verify the performance of our algorithm on multi-view data, we visualize the results on the MPV~\cite{MPV} dataset. As illustrated in Fig.~\ref{mpvvis}, the reference pose images are shown in the first row while the target garments are shown in the first column. The manipulated results are presented in other columns. It can be seen that our method captures the texture and pose well even with large variations in clothes design and viewpoint change. Besides, it generates realistic body parts even if they are unseen from the reference images which demonstrates the robustness of our method.

\begin{figure}[t]
\centering
\includegraphics[width= 12cm]{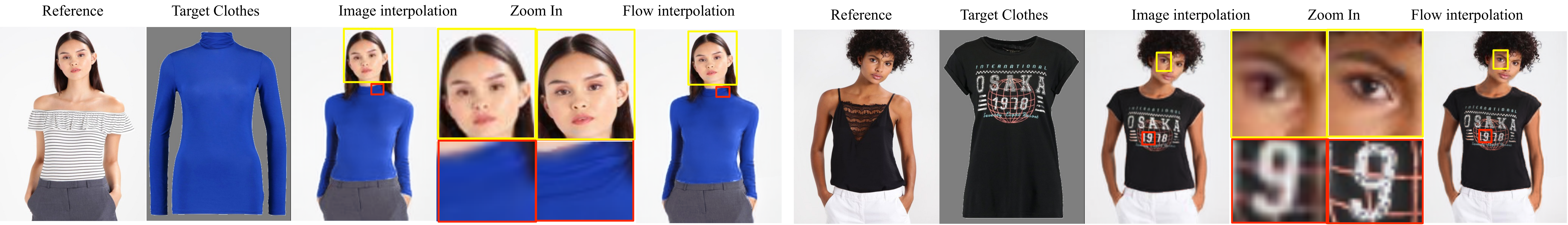}

   \caption{Comparisons with image interpolation and flow interpolation. Flow interpolation has the advantage of scaling to the higher resolution. \label{test_high}}

\end{figure}

\textbf{Inference at higher resolutions.}
In contrast to previous methods that can only predict at the inference stage the fixed size of images which is the same as training, our method is able to perform try-on at a higher resolution without re-training. We apply the model trained at $256 \times 192$ resolution to test the images with the size of $512 \times 384$ from VITON-HD~\cite{han2018viton}. It is obvious to be seen in Fig.~\ref{test_high} that our pure flow-based approach can retain finer texture information from the original image by linear interpolating the flows, as compared with the image-level interpolation scheme. It helps preserve the clarity of the characters and the patterns of the clothes, which verifies the scalability of our model.

\begin{table}[t]
\caption{Comparisons with State-of-the-art methods on VITON under paired setting.}
\scalebox{0.8}{ 
\begin{tabularx}{15cm}{p{1.6cm}|X<{\centering} X<{\centering} X<{\centering} X<{\centering}X<{\centering}}
\toprule[1.5pt]
Methods &CP-VTON \cite{CPVTON}&ClothFlow \cite{han2019clothflow}&ACGPN \cite{AGCPN}&ZFlow \cite{zflow}& \textbf{SDAFN(\tiny{Ours})}\\
\midrule[1pt]
FID ($\downarrow$)  &30.50&23.68&-&15.17& \textbf{10.97}\\
SSIM ($\uparrow$)  &0.784&0.843&0.845&0.885 & \textbf{0.888}\\
IS ($\uparrow$)  &2.757&-&2.829&- & \textbf{2.859}\\
PSNR ($\uparrow$)  &21.01&23.60&-&25.46 & \textbf{26.48}\\
\bottomrule[1.5pt]
\end{tabularx}}
\label{pair}
\end{table}

\begin{table*}[t]
\caption{Comparisons with state-of-the-art methods on VITON and MPV datasets under unpaired setting. For FID, the lower is the better.}
\scalebox{0.8}{ 
\begin{tabularx}{15cm}{p{2.4cm}|X<{\centering} X<{\centering}  X<{\centering} X<{\centering} X<{\centering} X<{\centering} X<{\centering}}
\toprule[1.5pt]
Methods &CP-VTON \cite{CPVTON}&ClothFlow \cite{han2019clothflow}&ACGPN \cite{AGCPN}&SDAFN (Ours)&WUTON \cite{issenhuth2020not}&PF-APN \cite{ge2021parser}& \textbf{SDAFN+ (Ours)}\\
\midrule[1pt]
VITON (\small{FID$\downarrow$}) &24.43&14.43&15.67&12.05&-&10.09 & \textbf{9.46}\\
\midrule[1pt]
MPV \ \ \ (\small{FID$\downarrow$})  &-&-&-&8.245&7.927 & 6.429 &\textbf{5.805} \\
\bottomrule[1.5pt]
\end{tabularx}}
\label{unpair}
\end{table*}

\subsection{Quantitative Results.} 
We conduct both paired and unpaired settings to quantitatively compare our work with baseline methods including CP-VTON \cite{CPVTON}, Clothflow\cite{han2019clothflow}, ACGPN\cite{AGCPN}, PFAFN~\cite{ge2021parser}, WUTON~\cite{issenhuth2020not} and ZFlow\cite{zflow}

\textbf{Paired setting.} In the paired setting, we use the Structure Similarity Index Measure (SSIM)~\cite{seshadrinathan2008unifying}, the Peak Signal-to-Noise Ratio (PNSR)~\cite{hore2010image} and the Fréchet Inception Distance(FID)~\cite{heusel2017gans} to measure the similarity between the synthesized image and ground truth image. The Inception Score (IS)~\cite{salimans2016improved} is applied to evaluate the realism of the generated images. We take the target image (the same person wearing the same clothes) as the ground truth images which are sued to compare with the synthesized image for computing these metrics. It is noted that PFAFN and WUTON were removed from those measurements as they need to take the target image as input for inference.

As shown in Table~\ref{pair}, our approach consistently outperforms the baselines methods under all the metrics. Achieving the best FID and IS scores shows our ability to obtain better visual quality in results and higher SSIM and PSNR scores demonstrate our advantage of accuracy.

\textbf{Unpaired setting.} Under the unpaired setting, there is no ground truth image for comparison. We directly adopt FID~\cite{heusel2017gans} to evaluate the similarity between the generated images and the real images. As reported in Table~\ref{unpair}, we compare both parser-free and parser-based methods. In particular, for a fair comparison, we train SDAFN+ model like PFAPN~\cite{ge2021parser} to compare, which allows us to directly input the original image without pose keypoints. In detail, SDAFN+ applies the prediction of SDAFN to replace the masked person and pose keypoints as the input for training. Compared with parser-based methods, we achieve obvious
improvements without segmentation guidance in FID. Compared with the parser-free method, the better FID on both VITON and MPV datasets indicates the generality of our method.

\begin{figure}[t]
\centering
\includegraphics[width= 10cm]{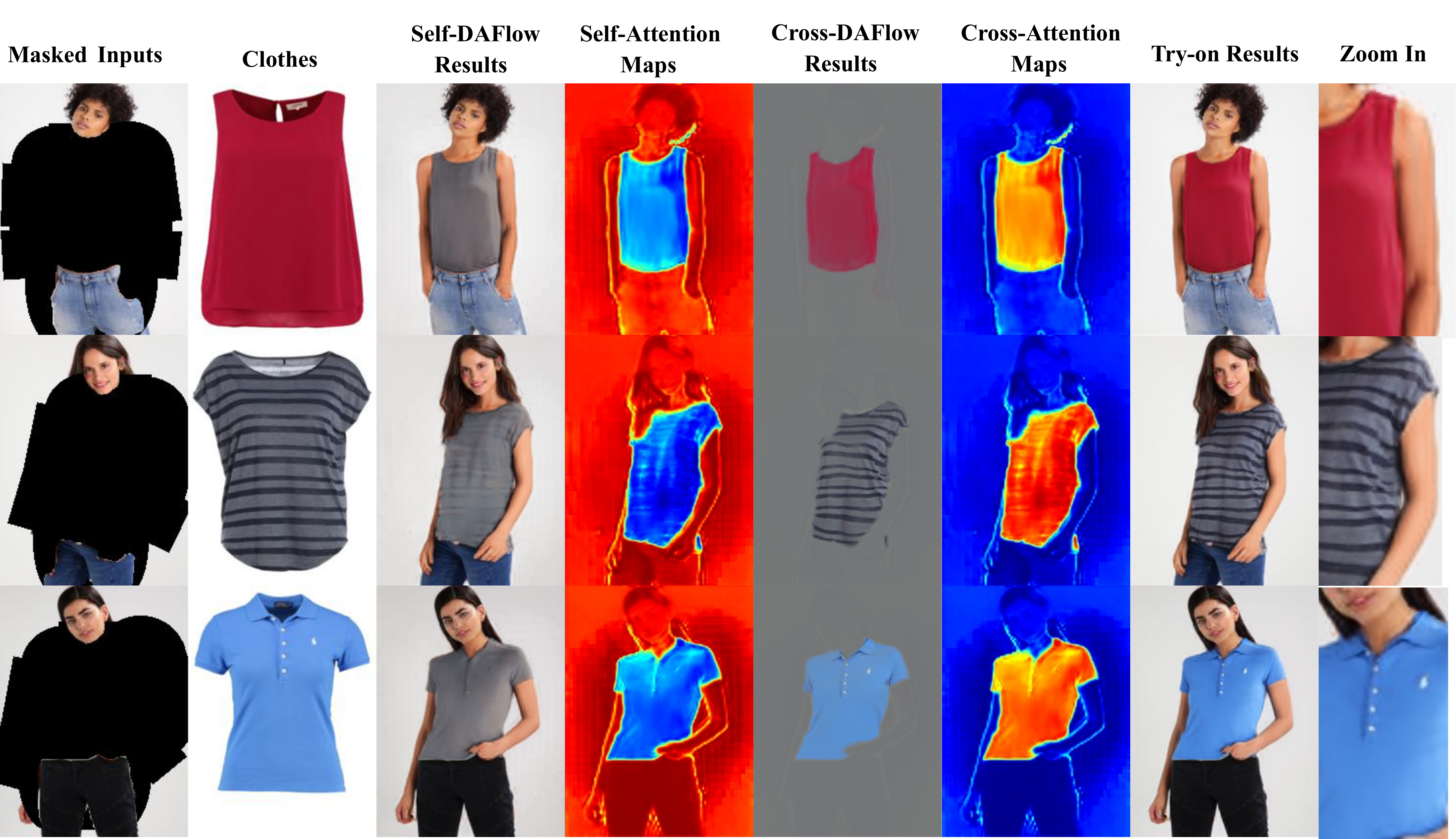}
   \caption{Visualization of the attention maps and results of deformable attention warping. \label{att}}

\end{figure}

\subsection{Ablation Study.}
In this section, we mainly evaluate the effectiveness of the proposed deformable attention flow module.

\textbf{Deformable Attention Flow.} 
The choice of the sampling key number $K$ in deformable attention flow is studied. Setting $K$ to 1 can be regarded as the traditional flow operation. With the increase of $K$, the performance gradually improves and drops slightly after it reaches 6, as shown in Table \ref{tab:tb1}. It shows that sampling multiple attention regions can effectively improve the flow operation. We set it to be 6 as a balance between performance and computational cost. In addition, as $K$ increases, more GPU memory is consumed for the sampling and warp operations but increases limited time consumption.

\begin{table*}[t]
\scalebox{1}{ 
\begin{floatrow}
\capbtabbox{
\begin{tabularx}{5cm}{p{0.7cm}| X<{\centering}X<{\centering}}
\toprule[1.5pt]

K &SSIM&PSNR\\
\midrule[1pt]
1 &0.833&22.43\\
2 &0.843&23.25\\
4 &0.868&24.71\\
6 &\textbf{0.888}&\textbf{26.48}\\
8 &0.878&25.35\\
\bottomrule[1.5pt]
\end{tabularx}
}{
 \caption{Performances of models with different sampling number.}
 \label{tab:tb1}
}
\capbtabbox{
 \begin{tabularx}{5cm}{p{2.3cm}|X<{\centering}X<{\centering}}
\toprule[1.5pt]
Config &SSIM&PSNR\\
\midrule[1pt]
Baseline &0.818&20.88\\
+Cascade &0.827&21.83\\
+Shallow En. &0.833&22.43\\
+DAFN& \textbf{0.888}& \textbf{26.48}\\
\bottomrule[1.5pt]
\end{tabularx}
}{
 \caption{The effect of different modules.}
 \label{tab:tb2}
}
\end{floatrow}}

\end{table*}

\textbf{Visualization of Attention.} In Fig.~\ref{att}, we visualize self- and cross-DAFlow along with the intermediate attention maps. After the self-DAFlow process, the clear clothes and torso shape are generated along with shadows which renders the realistic human body. The cross-DAFlow not only predicts accurate warping results but also retrains the fine textural details of the clothes, even with the folds on the t-shirt as shown in the second row. The attention maps show strong evidence that our flow module is able to learn the 3D priors where the clothes fit the skin and make the final try-on more realistic. In general, our method effectively decouples the textural and structural information and learns to add 3D shadow effect without introducing auxiliary models or labels.

\textbf{Modular Ablation Study.} 
The ablation study on the effectiveness of each module in our framework is reported in Table.\ref{tab:tb2}. It is shown that both SSIM and PSNR metrics increase with each of the modules added, including cascaded flow estimation, shallow encoder and decoder, and deformable attention flow network (DAFN). Among them, DAFN contributes the most to the improvement. The best performance is reported when integrating all the modules together.

\section{Applications on Other Tasks}
In this section, we mainly verify the versatility of the proposed deformable attention flow on two other image editing tasks, namely multi-view synthesis and image animation. To deal with only one image transformation (compared to paired images given in try-on task), we remove the self-MFE in DAFN  from our model. More details for implementation can be found in the supplementary.

\begin{figure}[t]
\centering
\includegraphics[width= 10cm]{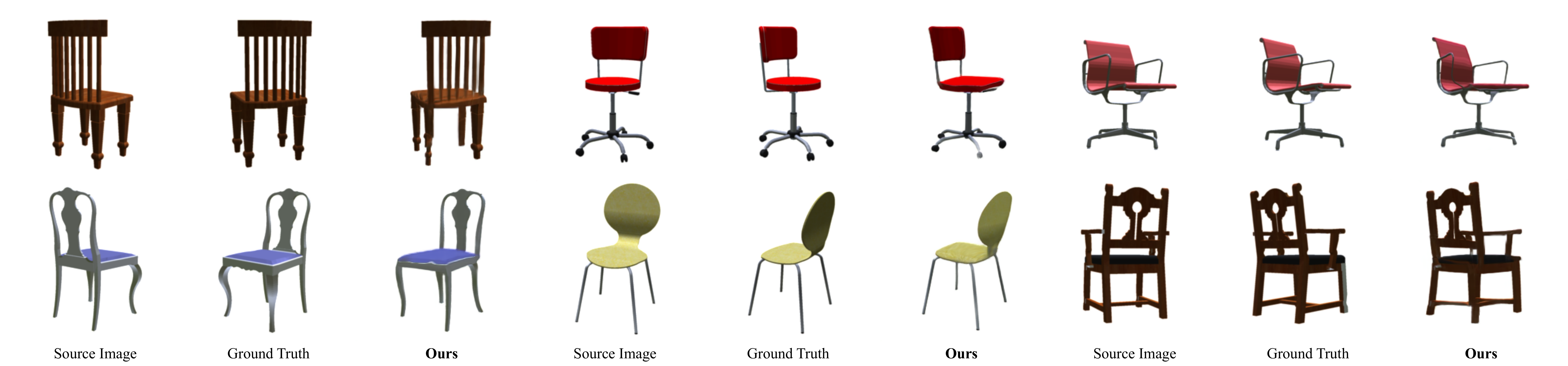}

   \caption{Qualitative results of the multi-view synthesis on ShapeNet.\label{shape}}

\end{figure}

\textbf{Multi-view synthesis.}
View synthesis aims to generate a novel view of an object given one single image as input. We apply our proposed SDAFN to learn the structural correlations of the same object under different viewing. We conduct experiments on the ShapeNet~\cite{chang2015shapenet} chairs dataset. As demonstrated in Fig.~\ref{shape}, our method is able to predict the one-shot input chair under different views with accurate textures. It successfully reconstructs the unseen parts which are under occlusion from the input image.

\begin{figure}[t]
\centering
\includegraphics[width= 10cm]{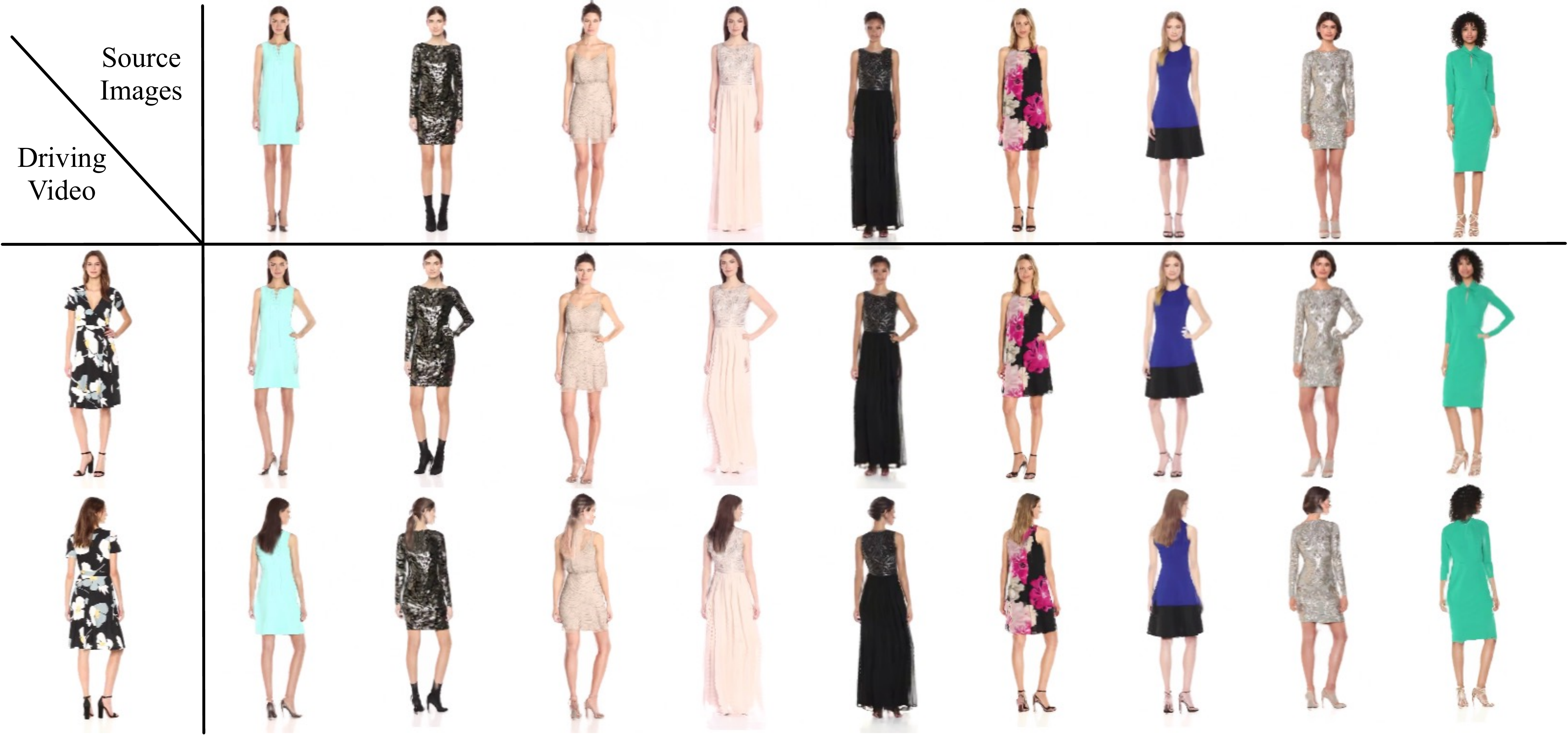}

   \caption{Qualitative results of the image animation task on FashionVideo.\label{fashionv}}

\end{figure}

\textbf{Image animation.}
Given an input image and a driving video, image animation is to generate a video sequence so that the object in the source image is animated according to the motion of the driving video. We experiment on the Fashion Video dataset~\cite{zablotskaia2019dwnet}, using 500 videos for training and 100 videos for testing. Example animations produced by our SDAFN on two actions are shown in Fig.\ref{fashionv}, where accurate reconstruction of the input pose is generated even in the case of complex motion like the back of the body.

\section{Conclusions}
In this paper, we present a novel single-stage virtual try-on framework. With only pose map as guidance, our model generates photo-realistic fitting results in an end-to-end manner. The proposed deformable attention flow (DAFlow) module estimates the accurate structure while retaining the vivid texture. It synthesizes photo-realistic human torso and fitting clothes with 3D shadows. Extensive experiments and evaluations show that our proposed method not only achieves superior performance on virtual try-on, but also can be extended to other image editing tasks, such as multi-view synthesis and image animation.
\clearpage
%
%
\bibliographystyle{splncs04}
\bibliography{egbib}
\end{document}